\documentclass[a4paper]{article}

\usepackage{itgspeech2023}    %
\usepackage{times}            %
\usepackage[english]{babel}   %
\usepackage[utf8]{inputenc}%
\usepackage[T1]{fontenc}      %
\usepackage[sort&compress,numbers]{natbib}	%
\usepackage{amsmath,amssymb}
\usepackage{graphicx}
\usepackage{microtype}
\usepackage{setspace}
\usepackage[colorlinks=false,pdfborder={0 0 0}]{hyperref}
\usepackage[usenames,dvipsnames]{xcolor}
\usepackage{multirow}
\usepackage{float}
\usepackage[symbol,bottom]{footmisc}
\usepackage{siunitx}
\usepackage{xspace}
\usepackage{adjustbox}
\usepackage[normalem]{ulem}
\usepackage{wrapfig}
\usepackage{enumitem}
\usepackage[acronym,toc,shortcuts,nonumberlist]{glossaries}

\newacronym{AM}{AM}{acoustic model}
\newacronym{AMI}{AMI}{Augmented Multi-party Interaction}
\newacronym{ARQ}{ARQ}{automatic repeat request}
\newacronym{ASR}{ASR}{automatic speech recognition}
\newacronym{AvL}{AvL}{Adversarial Learning}
\newacronym[longplural={bi-directional long-short term memories}]{BLSTM}{BLSTM}{bi-directional long-short term memory}
\newacronym{BSS}{BSS}{blind speech separation}
\newacronym{CART}{CART}{classification and regression tree}
\newacronym{CE}{CE}{cross entropy}
\newacronym{CER}{CER}{character error rate}
\newacronym{CDp}{CDp}{context dependent phoneme}
\newacronym{CMLLR}{CMLLR}{constrained maximum likelihood linear regression}
\newacronym{CNN}{CNN}{convolutional neural network}
\newacronym{CPC}{CPC}{contrastive predictive coding}
\newacronym{CTC}{CTC}{connectionist temporal classification}
\newacronym{DCT}{DCT}{discrete cosine transform}
\newacronym{DL}{DL}{deep learning}
\newacronym{DNN}{DNN}{deep neural network}
\newacronym{DNN-HMM}{DNN-HMM}{deep neural network hidden Markov model}
\newacronym{ELBO}{ELBO}{evidence lower bound}
\newacronym{EM}{EM}{expectation maximization}
\newacronym{fCE}{fCE}{frame-wise cross-entropy}
\newacronym{FE}{FE}{feature extractor}
\newacronym{FFNN}{FFNN}{feed-forward neural network}
\newacronym{FIR}{FIR}{finite impulse response}
\newacronym{FMLLR}{fMLLR}{Feature space Maximum Likelihood Linear Regression}
\newacronym{G2P}{G2P}{grapheme-to-phoneme conversion}
\newacronym{GAN}{GAN}{generative adversarial network}
\newacronym{GMM}{GMM}{Gaussian mixture model}
\newacronym{GMM-HMM}{GMM-HMM}{Gaussian mixture model hidden Markov model}
\newacronym{GPU}{GPU}{graphics processing unit}
\newacronym{HMM}{HMM}{hidden Markov model}
\newacronym{IHM}{IHM}{individual headset microphones}
\newacronym{IIR}{IIR}{infinite impulse response}
\newacronym{LAS}{LAS}{listen-attend-spell}
\newacronym{LC-BLSTM}{LC-BLSTM}{latency-controlled bidirectional long-short term memory}
\newacronym{LDA}{LDA}{linear discriminant analysis}
\newacronym{LM}{LM}{language model}
\newacronym{LR}{LR}{learning rate}
\newacronym[longplural={long-short term memories}]{LSTM}{LSTM}{long-short term memory}
\newacronym{MDM}{MDM}{multiple distant microphones}
\newacronym{MFA}{MFA}{Multi-scale Feature Aggregation}
\newacronym{MFCC}{MFCC}{Mel-frequency cepstral coefficients}
\newacronym{MHSA}{MHSA}{multi-head self-attention}
\newacronym{MLLR}{MLLR}{Maximum Likelihood Linear Regression}
\newacronym{MRES}{MRES}{multi-resolutional}
\newacronym{MSE}{MSE}{mean squared error}
\newacronym{MT}{MT}{machine translation}
\newacronym{MTL}{MTL}{Multi-task Learning}
\newacronym{NLP}{NLP}{natural language processing}
\newacronym{NN}{NN}{neural network}
\newacronym{OCLR}{OCLR}{one cycle learning rate}
\newacronym{OOV}{OOV}{out-of-vocabulary}
\newacronym{PC2}{$\text{PC}^{\text{2}}$}{Paderborn Center for Parallel Computing}
\newacronym{RNA}{RNA}{recurrent neural aligner}
\newacronym{RNN}{RNN}{recurrent neural network}
\newacronym{RSAN}{RSAN}{recurrent selective attention network}
\newacronym{SAT}{SAT}{speaker adaptive training}
\newacronym{SDM}{SDM}{single distant microphone}
\newacronym{SDR}{SDR}{signal-to-distortion ratio}
\newacronym{SC}{SC}{supervised convolutional}
\newacronym{sMBR}{sMBR}{state-level minimum Bayes risk}
\newacronym{STFT}{STFT}{short time Fourier transform}
\newacronym{TDNN}{TDNN}{time delay neural network}
\newacronym[longplural={time-frequencies}]{tf}{tf}{time-frequency}
\newacronym{UBM}{UBM}{universal background model}
\newacronym{VAD}{VAD}{voice activity detection}
\newacronym{VTLN}{VTLN}{vocal tract length normalization}
\newacronym{cpWER}{cpWER}{concatenated minimum-permutation word error rate}
\newacronym{WER}{WER}{word error rate}
\newacronym{WERR}{WERR}{word error rate reduction}
\newacronym{WP}{WP}{work package}
\newacronym{WPE}{WPE}{weighted prediction error}
\newacronym{WSJ}{WSJ}{Wall Street Journal}

\newcommand{\tab}{Table~}
\newcommand{\fig}{Figure~}

\newcommand{\ivec}{i-vector\xspace}
\newcommand{\ivecs}{i-vectors\xspace}

\newcommand{\swb}{Switchboard\xspace}
\newcommand{\lbs}{LibriSpeech\xspace}
\newcommand{\devclean}{\textit{dev-clean}\xspace}
\newcommand{\devother}{\textit{dev-other}\xspace}
\newcommand{\testclean}{\textit{test-clean}\xspace}
\newcommand{\testother}{\textit{test-other}\xspace}
\newcommand{\transformer}{\textit{Transformer}\xspace}
\newcommand{\conformer}{\textit{Conformer}\xspace}

\newcommand{\xvec}{x-vector\xspace}
\newcommand{\xvecs}{x-vectors\xspace}
\newcommand{\cvec}{c-vector\xspace}

\newcommand{\weightedSimpleAdd}{Weighted-Simple-Add\xspace}

\newcommand{\avgpool}{\textit{average pooling}\xspace}
\newcommand{\statspool}{\textit{statistics pooling}\xspace}
\newcommand{\attpool}{\textit{attention-based pooling}\xspace}
\newcommand{\attstatspool}{\textit{attentive statistics pooling}\xspace}

\title{Analyzing And Improving Neural Speaker Embeddings for ASR}

\author{Christoph L\"uscher$^{*1,2}$, Jingjing Xu$^{*1,2}$, Mohammad Zeineldeen$^{1,2}$, Ralf Schl\"uter$^{1,2}$, Hermann Ney$^{1,2}$}

\address{$^1$Machine Learning and Human Language Technology, RWTH Aachen University, 52074 Aachen, Germany\\
$^2$AppTek GmbH, 52062 Aachen, Germany\\
    Email: \texttt{\{luescher,jxu,zeineldeen\}@cs.rwth-aachen.de}\\
    $^*$ Equal contribution
}

\begin{document}

\maketitle

\begin{abstract}
Neural speaker embeddings encode the speaker’s speech characteristics through a DNN model and are prevalent for speaker verification tasks.
However, only a few inconclusive studies have investigated the usage of neural speaker embeddings for an ASR system.
In this work, we present our efforts w.r.t integrating neural speaker embeddings into a \conformer-based hybrid HMM ASR system.
For ASR, our improved embedding extraction pipeline in combination with the \weightedSimpleAdd integration method results in \xvec and \cvec reaching on par performance with \ivecs.
We further analyze, compare and combine different speaker embeddings.
We improve our already strong baseline by switching to one cycle learning schedule while reducing the training time.
By further adding neural speaker embeddings, we gain additional improvements.
This results in our best \conformer-based hybrid ASR system with speaker embeddings achieving 9.0\% WER on Hub5’00 and Hub5’01 while only training on SWB 300h.
\end{abstract}

\glsresetall

\section{Introduction \& related work}
\label{sec:intro}

For quite some time, speaker adaptation methods have been used to build robust \gls{ASR} systems \cite{george2013ivec}, aiming at reducing the divergence of distribution caused by various speakers in the train and test datasets, compensating for differing vocal tracts, gender, accents, dialects, and further general speaker characteristics.
Methodically, speaker adaptation is subdivided into different types of approaches, namely model space and feature space approaches.

Model space approaches accommodate speaker-specific representations within the \gls{AM}
\cite{liao2012sat, dong2013sat, seide2011feature, swietojanski2014LHUC, swietojanski2016LHUC, gemello2006adaptation, li2010comparison}
or add additional auxiliary losses in a multitask or adversarial training fashion,
while feature space approaches focus on the input to the \gls{AM}.
For feature space approaches, two common methods exist. Firstly, transforming the input feature vectors to be speaker normalized or speaker-dependent, for example with Vocal Tract Length Normalization \cite{welling1999vtln} and Maximum Likelihood Linear Regression \cite{mllr,ghalehjegh2014cmllr}.
Secondly, by augmenting the model with speaker embeddings \cite{najim2011ivec,snyder2018xvector}.
The specific integration method for a chosen speaker embedding depends on the \gls{AM} architecture.
Concatenating the \ivec to the network input gives performance gains for a \gls{BLSTM} \gls{RNN} \gls{AM} in a hybrid modeling approach \cite{kitza2019cumulative}.
The identical integration method leads to performance degradation when utilizing a \conformer \gls{AM} \cite{deng2022confivec}.
In our previous work \cite{zeineldeen2022improving}, we proposed an integration method suited to the \conformer \gls{AM}: \weightedSimpleAdd, in which we add the weighted speaker embeddings to the input of the \gls{MHSA} module.
However, this approach only works well for \ivecs but not for neural speaker embeddings, in our case \xvecs.
This trend can also be observed throughout the research literature, as \ivecs are widely used for \gls{ASR} \cite{kitza2019cumulative, zeineldeen2022improving, george2013ivec, Zoltan2021conformer}, but \xvecs and other neural embeddings are not.
To the best of our knowledge, there has only been very limited research on neural speaker embeddings for \gls{ASR}.
Research on neural speaker embeddings has been focused on speaker verification and recognition.
For these two tasks, neural speaker embeddings perform highly competitive.
In \cite{rownicka2019embeddings,rouhe2020speaker}, (neural) speaker embeddings have been applied to different \gls{ASR} systems, but no clear statement on which neural or non-neural speaker embedding to favour can be made.
Besides, \cite{rownicka2019embeddings,rouhe2020speaker} have the drawback that the models utilized are not state-of-the-art anymore, namely \gls{DNN} or \gls{BLSTM} based.
In this work, we focus on understanding the short-comings of current neural speaker embeddings for \gls{ASR}.
Additionally, we integrate the neural speaker embeddings into a highly competitive \conformer \gls{AM} baseline.
The inherently more powerful neural model compared to the \gls{GMM} should extract more expressive speaker embeddings.
But the speaker embedding extraction is tailored to the speaker verification and recognition task,
thus leading to performance degradation when applied to \gls{ASR}.
We investigate methods to extract speaker embeddings more suitable for the \gls{ASR} task.\\
The main contributions of this paper are:
\begin{enumerate}
    \item proposing an improved extraction pipeline for neural speaker embeddings, which are performant in an \gls{ASR} system. The resulting \xvec and \cvec from our improved extraction pipeline improve our \gls{ASR} system by $\sim$3\% relative \gls{WER} on Hub5'00.
    \item improving our \swb baseline by using \gls{OCLR}, leading to a relative \gls{WER} improvement of $\sim$3\% relative and a 17\% reduction in overall training time.
    \item verifying the \weightedSimpleAdd method on the \lbs dataset, resulting in a relative \gls{WER} reduction of 5.6\% on \testother.
\end{enumerate}
We introduce the combination of these methods and show that neural speaker embeddings can reach the same performance as \gls{GMM}-based speaker embeddings, namely \ivecs.
Although the individual parts of the pipeline are not new, the application in this manner for \gls{ASR} is novel.
Additionally, we demonstrate a training setup in which we can achieve the same performance as \ivecs.
Switching from a \gls{GMM}-based speaker embedding to a neural-based one without performance degradation,
opens up the possibility of integrating the speaker embedding network into the neural \gls{AM} encoder, leading to joint and multi-task training strategies.

\section{Speaker embeddings}
\label{sec:methods}

Every speaker's speech has unique characteristics due to gender, age, vocal tract variations, personal speaking style, intonation, pronunciation pattern, etc.
Speaker embeddings capture these speaker characteristics from the speech signal in the form of a learnable low-dimensional fixed-size vector.
The classical technique is based on a \gls{GMM}-\gls{UBM} system which extracts an \ivec \cite{najim2011ivec} as a feature vector.

\subsection{Neural speaker embedding}
Neural speaker embeddings are extracted from the bottleneck layer of a \gls{DNN},
which given speech data as input is trained in a supervised manner to discriminate between speakers.
Given the strong representation capability of \glspl{DNN} in learning highly abstract
features, neural embeddings have outperformed \ivecs in speaker verification
and speaker identification tasks \cite{snyder2018xvector,Jung2019rvector,zhang2022cvec}.
However, when applying neural speaker embeddings as a speaker adaptation method for \gls{ASR},
in general, \ivecs still outperform neural speaker embeddings \cite{zeineldeen2022improving,rouhe2020speaker}.
In this work, we propose an improved extraction pipeline for neural speaker
embeddings which is more suitable for speaker adaptation in ASR tasks.

\subsubsection{Neural extraction model}
Choosing an appropriate \gls{NN} for the neural speaker embeddings is vital.
In this work, two neural architectures are chosen: the \gls{TDNN} based \xvec \cite{snyder2018xvector}
and the \gls{MFA}-\conformer based \cvec \cite{zhang2022cvec}.
The well-known \xvec captures the speaker characteristics locally via the \gls{TDNN} structures in the bottom layers.
The newly proposed \gls{MFA}-\conformer model captures speaker characteristics locally AND globally via a
self-attention mechanism and aggregates hidden representations from multiple layers within the \gls{NN}.
Since the lower layers are more significant for learning speaker discriminative information \cite{tang2019multilevel},
such multi-level aggregation can make the speaker representation more robust.
Both models apply a temporal pooling layer to aggregate frame-level speaker features to obtain an utterance-level speaker embedding.
This aggregation across the time dimension is crucial for extracting speaker embeddings for \gls{ASR}.
In this work, we compare four different temporal pooling methods within the \gls{NN}: \avgpool, \statspool \cite{snyder2017statpooling}, \attpool \cite{rahman2018attention}, and \attstatspool \cite{koji2018attentive}.

\subsubsection{Post-processing methods}
After extracting the embeddings from the \gls{NN}, post-pro\-cessing is applied to boost the quality of the speaker representations.
We apply three standard post-processing procedures.
Firstly, subtract the global mean to center the representations.
Secondly, use \gls{LDA} to maximize the variance between the speaker embeddings of different speakers clusters and
minimizes intra-speaker variance caused by channel effects \cite{najim2011ivec}.
Thirdly, apply length (L2) normalization to normalize the euclidean length of each embedding to unit length \cite{rouhe2020speaker}.
Besides, in order to generate embeddings on a recording- or speaker-level, we simply average the embeddings corresponding to the same recording or speaker.

\subsubsection{Improved extraction pipeline}
An \ivec is designed to capture the highest mode from the total variability space,
and can therefore encode both speaker and channel characteristics.
However, the neural speaker embedding is trained to discriminate between speakers
and thus focuses on speaker-specific information.
As a result, the neural speaker embeddings capture less information, which might be helpful for speaker adaptation,
namely channel and other acoustic characteristics, compared to the \ivec.
This information loss could be a cause of the performance degradation of neural speaker
embeddings \cite{rownicka2019embeddings, zeineldeen2022improving}.

\begin{figure}[t]
  \centering
  \includegraphics[width=8.217cm]{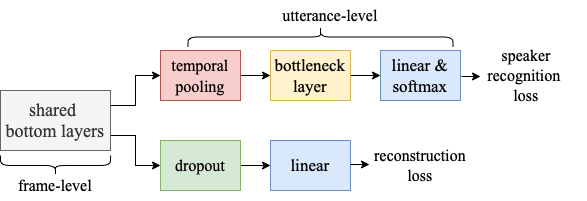}
  \caption{\textit{The speaker embedding model with auxiliary reconstruction loss.}}
  \label{fig:spk_model_with_reconstruction_loss}
\end{figure}

To let the neural speaker embeddings encode additional channel and acoustic information, on the layer below the temporal pooling layer, we add an auxiliary reconstruction loss.
It consists of a linear layer with dropout on the input, followed by an auxiliary \gls{MSE} loss to reconstruct the acoustic inputs, which can be seen in \fig\ref{fig:spk_model_with_reconstruction_loss}.
Suppose the layer representation is $h_t$ and the acoustic inputs are $x_t$.
The reconstruction loss can be formulated as
\[
\mathcal{L} =\frac{1}{2}\sum_{t=1}^{T}\sum_{c=1}^{C}(h_{t,c}-x_{t,c})^2
\]
where C is the corresponding feature dimension.
This allows us to capture additional channel and acoustic information within the speaker embedding.
Overall, we propose a better extraction pipeline: 1. train speaker embedding extractor with additional reconstruction loss 2. extract speaker embeddings from bottleneck layer 3. average over recordings 4. subtract the global mean

\section{Experimental setup}
\label{sec:setup}

The experiments are conducted on \swb 300h~\cite{Godfrey1992SWB} and \lbs 960h \cite{libri_dataset} datasets.
For \swb 300h dataset, we use Hub5'00 as the development set which consists of Switchboard (SWB) and CallHome (CH) parts.
CH part is much noisier.
We use Hub5'01 as test set.
For \lbs, we use the \devclean and \devother as the development set.
As the test set, we use \testclean and \testother.
The RETURNN \cite{zeyer2018:returnn} toolkit is used to train the acoustic models and
RASR \cite{wiesler2014rasr} toolkit is used for decoding.
All our configuration files and code to reproduce our results can be found
online\footnote{\scriptsize\url{https://bit.ly/train_pipeline}}.

\subsection{Baseline}
We use the same \conformer architecture and experimental setting as in our previous work \cite{zeineldeen2022improving}.
However, we replace the newbob \gls{LR} schedule with \gls{OCLR} schedule \cite{smith2019super}.
\gls{OCLR} can improve neural network training time without hurting performance, i.e. so-called super-convergence phenomenon.
Our \gls{OCLR} schedule consists of three phases, in the same manner as in \cite{zhou2022efficient}.
Firstly, the \gls{LR} linearly increases from $2e-3$ to $2e-2$ for 16 full epochs.
Secondly, the \gls{LR} linearly decreases from $2e-2$ to $2e-3$ for another 16 full epochs.
Finally, 10 more epochs are used to further decrease the learning rate to $1e-7$.
For \lbs, we apply the same training recipe with two changes.
First, we increase the dimension of the feed-forward module to 2048 and the attention dimension to 512 with 8 heads.
Second, we extend the \gls{LR} schedule to fully utilize the larger amount of training data.

We use the lattice-based version of \glsentrylong{sMBR} criterion \cite{Gibson2006HypothesisSF}.
The lattices are generated using the best \conformer \gls{AM} and a bigram \gls{LM}.
A small constant learning rate of 4e-5 is used.

We use 4-gram count-based \gls{LM} and \gls{LSTM} \GLS{LM} in first pass decoding \cite{beck2019lstm} and \transformer \gls{LM} for lattice rescoring.
For \swb, the \gls{LSTM} and \transformer \glspl{LM} have perplexity 51.3 and 48.1 on Hub5'00 respectively.
For \lbs, we use the official 4-gram count-based \gls{LM} and a \gls{LSTM} \gls{LM} in first pass decoding, more details can be found in \cite{Lscher2019RWTHAS}.
If not further specified, we use the count-based \gls{LM}.

\subsection{Speaker embeddings extraction}
All speaker embeddings are trained on either \swb 300h or \lbs 960h dataset.
The total number of speakers is 520 or 2338, accordingly.
Our \ivec extraction pipeline follows the recipe described in \cite{kitza2019cumulative}.
Empirically, we observe that 200-dim \ivecs works best for us while experimenting with 100-dim, 200-dim and 300-dim.
Our \xvec TDNN system follows the same architecture described in \cite{snyder2018xvector}.
The structure of the c-vector \gls{MFA}-\conformer framework is based on \cite{zhang2022cvec}.
We use 6 conformer blocks for the embedding extractor.
The attention dimension of each \gls{MHSA} module is 384 with 6 attention heads.
The dimension of the feed-forward module is set to 1536.
The bottleneck feature dimension is set to 512 for both embedding models.
For training the neural speaker embeddings, we split the training data into a train and cross-validation set.
The speaker identification is reported on a cross-validation set.

\section{Results}
\label{sec:results}

\subsection{Improved baseline}
In \tab\ref{tab:baseline_oclr}, we present our results using different \gls{LR} schedules and apply our \weightedSimpleAdd method on the \swb dataset.
We observe that switching from newbob \gls{LR} to \gls{OCLR} schedule greatly enhances training effectiveness,
allowing a reduction from 50 training epochs to 43,
while improving \gls{WER} from 10.7\% to 10.4\% on Hub5’00.
One possible explanation of this phenomenon is that a larger learning rate helps regularize the training.
We also show that integrating \ivecs using {\weightedSimpleAdd} outperforms
basic concatenation of the \ivec to the model input and improves WER from 10.4\% to 10.1\% on Hub5'00.
A comparison experiment using the same hyper-parameters but without i-vector integration proves that the
improvement does not stem from simply training longer.

\begin{table}[ht]
  \centering
    \caption{\textit{WERs [\%] of improved baseline using one cycle learning rate schedule on Switchboard 300h dataset}}
    \label{tab:baseline_oclr}
    \setlength{\tabcolsep}{0.3em}
    \scalebox{0.98}{
    \begin{tabular}{|@{\hskip2pt}l@{\hskip0pt}|@{\hskip0pt}c@{\hskip0pt}|@{\hskip0pt}c@{\hskip0pt}|@{\hskip0pt}c@{\hskip0pt}|@{\hskip0pt}c@{\hskip0pt}|@{\hskip0pt}c@{\hskip0pt}|}
    \hline
    \multirow{3}{*}{LR schedule} & \multirow{3}{*}{\shortstack{I-vec \\ integration \\ method}} & \multicolumn{3}{c|}{WER [\%]} &
    \multirow{3}{*}{\shortstack{full\\epochs}} \\ \cline{3-5}
    & & \multicolumn{3}{c|}{Hub5'00} & \\ \cline{3-5}
    & & SWB & CH & Total & \\ \hline \hline
    newbob & - & 7.1 & 14.3 & 10.7 & 50 \\ \hline \hline
    OCLR & - & 6.8 & 13.9 & 10.4 & 43 \\ \hline
    + long train & - & 6.6 & 14.1 & 10.4 & \multirow{1}{*}{68} \\
    \hline
    \multirow{2}{*}{+ i-vec} & append to input & 6.7 & 13.8 & 10.3 & \multirow{2}{*}{68} \\ \cline{2-5}
    & \weightedSimpleAdd & 6.7 & 13.5 & 10.1 & \\ \hline
    \end{tabular}
}
\end{table}

To show the transferability of the \weightedSimpleAdd method, we perform experiments on \lbs 960h dataset,
shown in \tab\ref{tab:ivec_on_librispeech}.
The results show around 6\% relative improvement in terms of WER on all sub-sets with the 4-gram LM.
However, the improvement gets smaller when we recognize with an LSTM LM.

\begin{table}[ht]
  \centering
  \caption{\textit{WERs [\%] speaker adaptive training by integrating i-vectors with Weighted-Simple-Add method on LibriSpeech 960h dataset}}
  \label{tab:ivec_on_librispeech}
  \setlength{\tabcolsep}{0.3em}
  \scalebox{0.98}{
  \begin{tabular}{|l|c|c|c|c|c|c|}
  \hline
  \multirow{2}{*}{Model} & \multirow{2}{*}{LM} & \multicolumn{2}{c|}{dev[\%]} & \multicolumn{2}{c|}{test[\%]} & \multirow{2}{*}{\shortstack{full\\epochs}} \\
  \cline{3-6}
  & & clean & other & clean & other & \\ \hline
  \multirow{2}{*}{baseline} & 4-gram & 2.9 & 6.7 & 3.2 & 7.1 & \multirow{2}{*}{15}\\
  \cline{2-6}
  & LSTM & 2.1 & 4.7 & 2.4 & 5.2 & \\ \hline
  + long train & 4-gram & 2.8 & 6.6 & 3.0 & 6.9 & 25 \\ \hline
  \multirow{2}{*}{+ i-vec} & 4-gram & 2.7 & 6.3 & 3.0 & 6.7 &  \multirow{2}{*}{25}\\
  \cline{2-6}
  & LSTM & 2.0 & 4.7 & 2.3 & 5.0 & \\
  \hline
\end{tabular}
}
\end{table}

\subsection{Temporal pooling comparison}
In order to extract speaker embeddings optimized for \gls{ASR},
we study the impacts of temporal pooling in embedding extractor layer.
The speaker embeddings are averaged over recordings and global mean is subtracted.

\begin{table}[ht]
  \centering
  \caption{\textit{Speaker identification accuracy [\%] of speaker embeddings extractor on dev set and WERs [\%] of SAT using x-vectors that are aggregated with different temporal pooling methods. }}
  \label{tab:temporal_pooling_comparison}
  \setlength{\tabcolsep}{0.2em}
  \scalebox{1}{\begin{tabular}{|c|c|c|c|c|}
  \hline
  \multirow{3}{*}{\shortstack{Temporal\\Pooling Method}} &
  \multirow{3}{*}{\shortstack{SpkId\\Accuracy [\%]}} &
  \multicolumn{3}{c|}{WER [\%]} \\ \cline{3-5}
  & & \multicolumn{3}{c|}{Hub5'00}\\ \cline{3-5}
  & & SWB & CH & Total \\ \hline
  average & 93.8 & 6.8 & 14.0 & 10.4 \\ \hline
  statistics & 89.9 & 6.7 & 14.0 & 10.3 \\ \hline
  attention-based & 94.3 & 6.8 & 14.0 & 10.4 \\ \hline
  attentive statistics & 90.5 & 6.6 & 13.7 & 10.2 \\ \hline
  \end{tabular}}
\end{table}

\tab\ref{tab:temporal_pooling_comparison} shows that the speaker identification accuracy
and \gls{ASR} performance, measured in \gls{WER}, do not correlate highly.
Both statistics pooling and attentive statistics pooling outperform average pooling and attention-based pooling
respectively regarding \gls{WER}, but show weaker performance for speaker identification accuracy.
This indicates that including the standard deviation gives some additional useful information for \gls{ASR} task.

\subsection{Neural speaker embedding post processing}
In \tab\ref{tab:post_processing_cmp}, we compare methods for improving the neural speaker embeddings for \gls{ASR}.

\begin{table}[ht]
    \centering
    \caption{\textit{WERs [\%] of SAT using DNN-based speaker embeddings that are different in aspects of speaker recognizer model and post processing level.}}
    \scalebox{0.945}{\begin{tabular}{|@{\hskip0pt}c@{\hskip0pt}|@{\hskip0pt}c@{\hskip0pt}|@{\hskip0pt}c@{\hskip0pt}|@{\hskip0pt}c@{\hskip0pt}|@{\hskip0pt}c@{\hskip0pt}|@{\hskip0pt}c@{\hskip0pt}|@{\hskip0pt}c@{\hskip0pt}||@{\hskip0pt}c@{\hskip0pt}|@{\hskip0pt}c@{\hskip0pt}|@{\hskip0pt}c@{\hskip0pt}|}
    \hline
    \multirow{3}{*}{\shortstack{Post\\Process.\\Level}} &
    \multirow{3}{*}{\shortstack{Subtract\\mean}} &
    \multirow{3}{*}{\shortstack{LDA}} &
    \multirow{3}{*}{\shortstack{With\\reconst\\loss}} &
    \multicolumn{3}{@{\hskip0pt}c@{\hskip0pt}||}{X-vec} &  \multicolumn{3}{@{\hskip0pt}c@{\hskip0pt}|}{C-vec}\\ \cline{5-10}
    & & & & \multicolumn{3}{c||}{Hub5'00} & \multicolumn{3}{c|}{Hub5'00}\\ \cline{5-10}
    & & & & SWB & CH & Total & SWB & CH & Total \\ \hline \hline
    utterance & \multirow{3}{*}{no} & \multirow{4}{*}{no} & \multirow{5}{*}{no} & 6.9 & 14.1 & 10.5 & 6.7 & 14.1 & 10.4\\ \cline{1-1}\cline{5-10}
    recording & & & & \textbf{6.6} & 13.9 & 10.2 & 6.7 & 13.8 & 10.2 \\\cline{1-1}\cline{5-10}
    speaker & & & & 6.7 & 13.9 & 10.3 & \textbf{6.6} & 13.8 & 10.2 \\\cline{1-2}\cline{5-10}
   \multirow{3}{*}{recording} & \multirow{3}{*}{yes} & & & \textbf{6.6} & 13.7 & 10.2 & \textbf{6.6} & 13.8 & 10.2\\\cline{3-3}\cline{5-10}
   & & yes & & \textbf{6.6} & 13.8 & 10.2 & 6.7 & 13.7 & 10.2 \\ \cline{3-4} \cline{5-10}
   & & no & yes & 6.7 & \textbf{13.5} & \textbf{10.1} & \textbf{6.6} & 13.7 & \textbf{10.1} \\  \hline
  \end{tabular}}
    \label{tab:post_processing_cmp}
\end{table}

We observe that the \conformer \gls{AM} only improves with recording-wise or speaker-wise embeddings.
This could be due to the increased context embedded into the speaker embeddings.
Furthermore, we also notice that \xvec and \cvec have similar performance.
Subtracting the global mean gives no improvement to the overall Hub5'00 \gls{WER}, but only minor improvement in the subsets.
Applying LDA has almost no effect on performance.
The weighting factor for the reconstruction loss is set to 5.
With the reconstruction loss, the speaker identification accuracy of the TDNN embedding model would drop from 90.5\% to  85.8\%.
On the contrary, the WER of SAT improves from 10.2\% to 10.1\% on Hub5'00.

\subsection{Speaker embeddings comparison}
The comparison between \ivecs and neural speaker embeddings is reported in \tab\ref{tab:spk_embedding_cmp}.
With our proposed extraction pipeline, the \gls{WER} when integrating \xvecs is reduced by 3.8\% relative,
i.e., from 10.5\% to 10.1\%, reaching the same performance as with \ivec integration.
Combining \ivecs with neural embeddings by concatenation did not show any further improvement,
hinting that the different speaker embeddings contain the same information.
As a control experiment, we replaced the learned speaker embeddings with Gaussian noise to verify the concern that the speaker embeddings only had an effect due to a form of noise regularization.
The control experiments have the same performance as the baseline, showing that the \gls{AM}
utilizes the speaker embeddings in the correct way.

\begin{table}[ht]
  \centering
  \caption{\textit{WERs [\%] of SAT using different types of speaker embeddings.}}
  \label{tab:spk_embedding_cmp}
  \scalebox{1}{\begin{tabular}{|l|c|@{\hskip1pt}c@{\hskip1pt}|@{\hskip1pt}c@{\hskip1pt}|@{\hskip1pt}c@{\hskip1pt}|@{\hskip1pt}c@{\hskip1pt}|}
  \hline
  \multirow{3}{*}{\shortstack{Speaker embedding}} &
  \multirow{3}{*}{\shortstack{With\\reconst.\\loss}} &
  \multicolumn{4}{c|}{WER [\%]} \\ \cline{3-6}
  & & \multicolumn{3}{c|}{Hub5'00} & \multirow{2}{*}{\shortstack{Hub\\5'01}}\\ \cline{3-5}
  & & SWB & CH & Total & \\ \hline
  none & - & 6.8 & 13.9 & 10.4 & 10.7 \\ \hline \hline
  Gaussian noise & - & 6.7 & 14.1 & 10.4 & 10.6\\ \hline
  i-vec & - & 6.7 & \textbf{13.5} & \textbf{10.1} & 10.3 \\ \hline
  \multirow{2}{*}{x-vec} & no & 6.9 & 14.1 & 10.5 & 10.6 \\ \cline{2-6}
  & yes & 6.7 & 13.5 & \textbf{10.1} & 10.4\\ \hline
  \multirow{2}{*}{c-vec} & no & 6.7 & 14.1 & 10.4 & 10.6 \\ \cline{2-6}
  & yes & \textbf{6.6} & 13.7 & \textbf{10.1} & 10.5 \\ \hline
  i-vec + x-vec & yes & 6.7 & 13.6 & \textbf{10.1} & 10.3 \\ \hline
  i-vec + x-vec + c-vec & yes & \textbf{6.6} & 13.6 & \textbf{10.1} & 10.3 \\ \hline
\end{tabular}}
\end{table}

\subsection{Overall result}
In \tab\ref{overall_results_table}, we present a highly competitive and efficient \conformer hybrid \gls{ASR} system with 58M parameters and trained with 74 epochs.
We outperform the well-trained \conformer transducer system \cite{zhou2022efficient} and our previous work \cite{zeineldeen2022improving} with fewer epochs and shorter training time.
Our best \conformer \gls{ASR} system does not reach the state-of-the-art results by \cite{Zoltan2021conformer}.
However, two important methods applied in \cite{Zoltan2021conformer}, speed perturbation and a cross-utterance \gls{LM}, are not used in this work but could lead to further improvements.

\begin{table}[ht]
  \centering
  \caption{Overall WER [\%] comparison with literature.}
  \label{overall_results_table}
  \setlength{\tabcolsep}{0.2em}
  \scalebox{0.9}{
  \setlength\tabcolsep{1pt}
  \begin{tabular}{|c|c|c|c|c|c|c|c|c|c|}
    \hline
    \multirow{3}{*}{Work} & \multirow{3}{*}{\shortstack{ASR\\Arch.}}
    & \multirow{3}{*}{AM} & \multirow{3}{*}{LM} &  \multirow{3}{*}{\shortstack{seq\\disc}} &
    \multirow{3}{*}{\shortstack{ivec}} &
    \multirow{3}{*}{\shortstack{num\\param}} &
    \multirow{3}{*}{\shortstack{full\\epochs}} &
    \multicolumn{2}{c|}{WER [\%]} \\
    \cline{9-10}
    & & & & & & & & \multicolumn{2}{c|}{Hub} \\ \cline{9-10}
    & & & & & & & & 5'00 & 5'01 \\ \hline \hline

    \shortstack{\cite{zhou2022efficient}}
    & RNNT & Conf. & Trafo & yes & no & 75 & 86 & 9.2 & 9.3 \\ \hline \hline

    \multirow{1}{*}{\cite{Zoltan2021conformer}}
    & \multirow{1}{*}{LAS} & \multirow{1}{*}{Conf.} & Trafo &
    \multirow{1}{*}{no} & \multirow{1}{*}{yes} & \multirow{1}{*}{68} & \multirow{1}{*}{250} & 8.4 & 8.5 \\ \hline \hline

    \cite{hu2021bayesian} & \multirow{7}{*}{Hybrid} & TDNN & RNN & yes & yes & 19 & - &
    10.4 & - \\ \cline{1-1} \cline{3-10}

    \multirow{1}{*}{\shortstack{\cite{zeineldeen2022improving}}} & & \multirow{6}{*}{Conf.} & Trafo & \multirow{1}{*}{yes} & yes & \multirow{6}{*}{58} & 90 & 9.2 & 9.3 \\ \cline{1-1} \cline{4-6} \cline{8-10}

    \multirow{5}{*}{ours} & & & \multirow{3}{*}{4-gr} &
    \multirow{2}{*}{no} & no & & 43 & 10.4 & 10.7 \\ \cline{6-6} \cline{8-10}

    & & & &
    & \multirow{4}{*}{yes} & & 68 & 10.1 & 10.3 \\ \cline{5-5} \cline{8-10}

    & & & &
    \multirow{3}{*}{yes} & & & \multirow{3}{*}{74} & 10.0 & 10.1 \\ \cline{4-4} \cline{9-10}
    & & & LSTM & & & & & 9.2 & 9.3 \\ \cline{4-4} \cline{9-10}
    & & & Trafo & & & & & 9.0 & 9.0\\ \hline
  \end{tabular}
}
\end{table}

\section{Conclusion}
\label{sec:concl}

In this work, we focus our research on neural speaker embeddings for \gls{ASR}.
We begin our investigations by improving our strong baseline on the \swb dataset with the utilization of an \gls{OCLR} schedule.
We apply the \weightedSimpleAdd method on top of our \swb baseline and confirm the method on the \lbs dataset.
Focusing on the shortcomings of neural speaker embeddings compared to the conventional \ivecs, we propose an improved extraction pipeline.
Different design choices for the neural speaker embedding extraction pipeline are presented and compared.
With our proposed pipeline, the integration of \xvec or \cvec improves the \gls{ASR} system performance by $\sim$3\% relative, reaching on-par performance with \ivecs.
We think a more sophisticated reconstruction loss that captures the relevant information in the speaker embedding more accurately could push the performance of the neural speaker embedding \gls{ASR} system beyond the \gls{ASR} system with \ivecs.
Moreover, we tried to combine the \ivec with neural speaker embeddings but gained no further improvement.
This indicates that the neural speaker embeddings contain at least very similar information as the \ivecs.
The competitive performance of the neural speaker embeddings opens up possibilities for joint training and integrating neural speaker embeddings into the \gls{AM}.
Overall, we present a highly competitive and efficient \conformer hybrid \gls{ASR} system, approaching the state-of-the-art results but with a much smaller model and less training time.

\section{Acknowledgements}
\label{sec:ack}

\small
This work was partially supported by a Google Focused Award. The work reflects only the authors' views and none of the funding parties is responsible for any use that may be made of the information it contains.

\small
\bibliographystyle{ieeetr}
\bibliography{references}

\end{document}